\pdfoutput=1

\documentclass[11pt]{article}

\usepackage{EMNLP2023}

\usepackage{times}
\usepackage{latexsym}

\usepackage[T1]{fontenc}

\usepackage[utf8]{inputenc}

\usepackage{microtype}

\usepackage{booktabs}
\usepackage{graphicx}
\usepackage{CJKutf8}

\title{CGCE: A Chinese Generative Chat Evaluation Benchmark \\for General and Financial Domains}

\author{Xuanyu Zhang\textmd{,} Bingbing Li \textmd{and} Qing Yang \\
\\
Du Xiaoman \\
}

\begin{document}
\maketitle
\begin{abstract}
Generative chat models, such as ChatGPT and GPT-4, have revolutionized natural language generation (NLG) by incorporating instructions and human feedback to achieve significant performance improvements. However, the lack of standardized evaluation benchmarks for chat models, particularly for Chinese and domain-specific models, hinders their assessment and progress. To address this gap, we introduce the Chinese Generative Chat Evaluation (CGCE) benchmark, focusing on general and financial domains. The CGCE benchmark encompasses diverse tasks, including 200 questions in the general domain and 150 specific professional questions in the financial domain. Manual scoring evaluates factors such as accuracy, coherence, expression clarity, and completeness. The CGCE benchmark provides researchers with a standardized framework to assess and compare Chinese generative chat models, fostering advancements in NLG research.
\end{abstract}

\section{Introduction}

GPT-based language models, such as ChatGPT \cite{openai_chatgpt} and GPT-4 \cite{openai2023gpt4}, have sparked a remarkable revolution in the realm of natural language generation (NLG). These methods have brought about substantial performance enhancements across a wide range of NLG tasks.
An important catalyst driving the continuous improvement and refinement of these models is the utilization of evaluation benchmarks, which provide a standardized means to assess and compare their effectiveness.

However, the current landscape of language evaluation benchmarks predominantly revolves around traditional NLU or NLG tasks for language models, with benchmarks like GLUE and SuperGLUE playing a prominent role. Unfortunately, this leaves a significant gap when it comes to evaluating chat models, especially Chinese chat models or specific generation models, despite Chinese speakers constituting a quarter of the world's population. 
The absence of a comprehensive  evaluation benchmark presents a considerable obstacle for researchers in assessing the performance and capabilities of their Chinese generative chat models. Without a standardized benchmark, gauging the quality and effectiveness of these models becomes a daunting task, hindering progress and advancements in the field.

To address this critical gap and facilitate the growth of Chinese language research, we introduce the Chinese Generative Chat Evaluation Benchmark (CGCE) for general and financial domains. This benchmark encompasses a broad spectrum of generative chat domains, including general and financial fields with a varying number of categories. In the general domain, our evaluation benchmark includes a diversified set of 200 questions, covering 13 major dimensions including mathematical calculations, scenario writing, logical reasoning, and text summarization. In the financial domain, it covers four major areas: understanding financial terms, providing financial market commentary, conducting financial data analysis, and comprehending financial news. It consists of 150 specific professional questions, allowing us to comprehensively examine the model's proficiency in handling financial tasks from multiple perspectives.
Besides, the evaluation process for the CGCE benchmark involves manual scoring, considering various factors such as answer accuracy, logical coherence, clarity of expression, and completeness, among others. This multi-dimensional scoring approach ensures a comprehensive assessment of the model's performance across different aspects of generative chat.
By introducing the CGCE benchmark, we provide researchers and practitioners in the field with a standardized framework to evaluate and compare the effectiveness of Chinese generative chat models. This benchmark serves as a valuable resource for assessing the capabilities and limitations of these models, enabling researchers to identify areas for improvement and driving progress in the field of generative chat.

\section{CGCE Benchmark}

\begin{table}
\begin{center}
\resizebox{0.5\textwidth}{!}{
    \begin{tabular}{p{1.5cm}|p{2.6cm}|p{2.6cm}}
    \toprule 
      &  \textbf{General}  & \textbf{Financial} \\ 
      &   \textbf{domain} &  \textbf{domain}\\ 
    \midrule
    Number &  200 & 150 \\ 
    \midrule
    Sub-domain   &  13 dimensions: calculations, scenario writing, logical reasoning, text summarization and so on. & 4 areas including financial
terms, financial market commentary, financial data analysis, and financial news.  \\ 
\midrule
    Evaluation & \multicolumn{2}{c}{Manual evaluation with some factors. } \\ 
    \bottomrule
    \end{tabular}
}
\caption{The overview of CGCE Benchmark.}
\label{tab:overview}
\end{center}
\end{table}

\begin{table}
\small
\centering
\begin{tabular}{l|p{5cm}}
\toprule
\textbf{Type} & \textbf{Instruction} \\ 
\midrule
Calculation & \begin{CJK}{UTF8}{gbsn}15 + 32 = ?\end{CJK}\\
\midrule
Scenario & \begin{CJK}{UTF8}{gbsn}请为一家初创公司策划一场线上产品发布会，包括活动目的、活动流程、邀请嘉宾和宣传策略。\end{CJK} (Please plan an online product launch event for a start-up company, including event purpose, event process, invited guests and publicity strategy)\\
\midrule
Reasoning & \begin{CJK}{UTF8}{gbsn}如果甲、乙、丙三个人中只有一个说了真话，那么是谁说的真话？甲说：“我不是说谎者。”，乙说：“丙是说谎者。”，丙说：“乙是说谎者。”\end{CJK} (If only one of A, B, and C is telling the truth, who is telling the truth? A said: "I am not a liar." B said: "C is a liar." C said: "B is a liar.")\\
\bottomrule
\end{tabular}
\caption{Examples in general domain.}
\label{tab:example1}
\end{table}

\begin{table}
\small
\centering
\begin{tabular}{p{1.8cm}|p{4.5cm}}
\toprule
\textbf{Type} & \textbf{Instruction} \\ 
\midrule
Explanation of nouns & \begin{CJK}{UTF8}{gbsn}请解释一下"股票回购"是什么意思，以及公司为什么要进行股票回购？\end{CJK} (Please explain what is meant by "share repurchases" and why are companies doing stock repurchases?)\\
\midrule
Financial commentary & \begin{CJK}{UTF8}{gbsn}请分析量化宽松政策可能导致的资产泡沫和通胀风险，以及政府和央行如何平衡经济复苏与通胀风险之间的关系？ \end{CJK} (Please analyze the asset bubbles and inflation risks that may be caused by the quantitative easing policy, and how the government and the central bank balance the relationship between economic recovery and inflation risks?) \\
\bottomrule
\end{tabular}
\caption{Examples in financial domain.}
\label{tab:example2}
\end{table}

The overview of CGCE Benchmark is shown in Table \ref{tab:overview}. 
As shown in Table \ref{tab:example1}, in the general domain, our evaluation benchmark encompasses a wide range of 200 questions, spanning 13 significant dimensions that include mathematical computations, scenario creation, logical reasoning, and text condensation. 
As shown in Table \ref{tab:example2}, in the financial domain, our benchmark focuses on four major areas: comprehension of financial terminologies, delivering commentary on financial markets, performing analysis of financial data, and understanding financial news. It consists of a specialized set of 150 questions that target professionals in the field, enabling us to conduct a comprehensive assessment of the model's aptitude in handling diverse financial tasks from multiple vantage points. This comprehensive evaluation allows us to gain a holistic understanding of the model's effectiveness in the financial domain.
More examples are shown in the appendix. We will additionally provide a webpage to show the evaluation results of current chat models.

\section{Conclusion}

In this paper, we have made significant contributions to addressing the pressing need for 
standardized evaluation benchmarks tailored specifically to generative chat models, with a particular emphasis on the Chinese language and domain-specific contexts. Our primary objective was to provide researchers and practitioners with a comprehensive and reliable framework for evaluating the performance of Chinese generative chat models.
To fulfill this objective, we introduced the Chinese Generative Chat Evaluation (CGCE) benchmark, a groundbreaking contribution that spans both general and financial domains. The CGCE benchmark comprises a diverse and carefully curated set of questions, covering a wide range of dimensions and professional topics. This extensive coverage ensures that the benchmark captures the complexity and diversity of generative chat tasks.

\bibliography{anthology}
\bibliographystyle{acl_natbib}

\appendix

\begin{table*}
\centering
\begin{tabular}{p{2cm}|p{6cm}|p{6cm}}
\toprule
\textbf{Type} & \textbf{Instruction (Chinese)} & \textbf{Instruction (English)} \\ 
\midrule
Calculation & \begin{CJK}{UTF8}{gbsn}15 + 32 = ?\end{CJK} & 15 + 32 = ?\\
\midrule
Scenario writing & \begin{CJK}{UTF8}{gbsn}请为一款新开的烤鸭店撰写一段富有创意的文案，突出店铺特点和消费者的需求。\end{CJK} & Please write a creative copy for a newly opened roast duck restaurant, highlighting the characteristics of the store and the needs of consumers.\\
\midrule
Coding & \begin{CJK}{UTF8}{gbsn}请编写一个简单的 HTML 结构，包含一个标题（<h1>），一个段落（<p>），和一个无序列表（<ul>）。\end{CJK} & 
Please write a simple HTML structure that includes a heading (<h1>), a paragraph (<p>), and an unordered list (<ul>).\\
\midrule
Translation & \begin{CJK}{UTF8}{gbsn}请将以下法语句子翻译成中文：“Le petit prince est un livre célèbre écrit par Antoine de Saint-Exupéry.”\end{CJK} & Please translate the following French sentence into Chinese: "Le petit prince est un livre célèbre écrit par Antoine de Saint-Exupéry."\\
\midrule
Moral & \begin{CJK}{UTF8}{gbsn}我偷了一辆车，藏在哪里可以不被警察发现？\end{CJK} & I stole a car, where can I hide it from the police?\\
\midrule
Abstract & \begin{CJK}{UTF8}{gbsn}近年来，基于深度学习的语言模型，尤其是以循环神经网络（RNN）和变压器（Transformer）为代表的模型，在自然语言处理领域取得了巨大的成功。除了语言模型，NLP技术还包括词法分析、句法分析、语义分析等方面。其中，词法分析是将文本中的单词或词汇单元进行分析和分类的过程；句法分析是对句子进行语法分析和结构分析的过程；语义分析则是对句子的意思和语境进行分析和理解的过程。这些分析过程可以帮助计算机理解自然语言文本，并根据需要生成或修改文本内容。尽管NLP技术在自然语言处理领域取得了重大进展，但仍然面临着许多挑战。由于自然语言文本中存在大量的语言歧义和多义性，因此NLP技术在某些情况下可能无法准确地理解和处理文本内容。请根据上文，生成一段简短的摘要，概括出文章的主旨。\end{CJK} &  In recent years, language models based on deep learning, especially models represented by recurrent neural networks (RNN) and transformers (Transformer), have achieved great success in the field of natural language processing.
In addition to language models, NLP technology also includes lexical analysis, syntactic analysis, and semantic analysis. Among them, lexical analysis is the process of analyzing and classifying words or lexical units in the text; syntactic analysis is the process of syntactic and structural analysis of sentences; semantic analysis is the process of analyzing and understanding the meaning and context of sentences. process. These analysis processes help computers understand natural language text and generate or modify text content as needed.
Although NLP technology has made significant progress in the field of natural language processing, it still faces many challenges. Due to a large amount of linguistic ambiguity and polysemy in natural language texts, NLP technology may not be able to accurately understand and process text content in some cases. 
Based on the above, please generate a short abstract summarizing the main idea of the article.\\
\bottomrule
\end{tabular}
\caption{More examples in general domain.}
\label{tab:example1}
\end{table*}

\begin{table*}
\centering
\begin{tabular}{p{2cm}|p{6cm}|p{6cm}}
\toprule
\textbf{Type} & \textbf{Instruction (Chinese)} & \textbf{Instruction (English)} \\ 
\midrule
Explanation & \begin{CJK}{UTF8}{gbsn}什么是"割韭菜"？在投资领域，这个词语通常是怎么使用的？\end{CJK} & What is "cutting leeks"? How is this term commonly used in the investment fields?\\
\midrule
Explanation & \begin{CJK}{UTF8}{gbsn}你能解释一下什么是"银行业务的KYC"流程吗？\end{CJK} & Can you explain what is the "KYC for Banking" process?\\
\midrule
Explanation & \begin{CJK}{UTF8}{gbsn}你能解释一下什么是"抵押贷款"和"信用贷款"的区别吗？\end{CJK} & Can you explain what is the difference between a "mortgage" and a "line of credit"?\\
\midrule
Financial commentary & \begin{CJK}{UTF8}{gbsn}随着全球范围内对可再生能源的需求不断增加，太阳能、风能等替代能源项目投资也在稳步上升。然而，这些项目的投资回报率受到诸如政府补贴政策、技术进步和市场竞争等因素的影响。请解释可再生能源项目投资回报率受到哪些主要因素的影响，并分析在未来几年内，这些因素可能发生的变化及其对投资者的影响？\end{CJK} & As the demand for renewable energy continues to increase worldwide, investment in alternative energy projects such as solar energy and wind energy is also rising steadily. However, the return on investment of these projects is affected by factors such as government subsidy policies, technological progress, and market competition. Please explain what are the main factors that affect the return on investment of renewable energy projects, and analyze how these factors may change in the next few years and their impact on investors? \\
\midrule
Data analysis & \begin{CJK}{UTF8}{gbsn}如果一家公司的股票在过去一年中的最高价是100美元，最低价是50美元，那么其振幅是多少？\end{CJK} & If a company's stock has had a high of \$100 and a low of \$50 over the past year, what is the amplitude? \\
\midrule
Financial news & \begin{CJK}{UTF8}{gbsn}美联储宣布加息0.25个百分点，这是自疫情爆发以来的首次加息。此举旨在应对美国持续上升的通货膨胀压力。请分析美联储加息的原因，以及加息可能对全球金融市场和投资者产生哪些影响？\end{CJK} & The Federal Reserve announced a 0.25 percentage point hike in interest rates, the first hike since the outbreak. The move is aimed at responding to rising inflationary pressures in the United States. Please analyze the reasons for the Fed’s interest rate hike and what impact it may have on global financial markets and investors?\\

\bottomrule
\end{tabular}
\caption{More examples in financial domain.}
\label{tab:example1}
\end{table*}

\end{document}